\documentclass[11pt]{article}
\usepackage{amsmath}

\usepackage[preprint]{acl}

\usepackage{times}
\usepackage{latexsym}

\usepackage[T1]{fontenc}

\usepackage[utf8]{inputenc}

\usepackage{microtype}

\usepackage{inconsolata}
\usepackage{wrapfig}
\usepackage{graphicx}
\usepackage{hyperref}
\usepackage{url}
\usepackage{graphicx}
\usepackage{hyperref}
\usepackage{url}
\usepackage{verbatim}
\usepackage{amsthm}
\usepackage{wrapfig}
\usepackage{multirow}
\usepackage{booktabs}
\usepackage{array}
\usepackage{wrapfig}
\usepackage{bbm}
\usepackage{multicol}
\usepackage{amssymb}
\usepackage{wrapfig}
\usepackage{arydshln}
\usepackage[most]{tcolorbox} 
\definecolor{tsinghuapurple}{RGB}{102,8,116}
\newtcolorbox{alprompt}[1]{
        boxrule = 1pt,
        fontupper = \small\tt,
        fonttitle = \bf\color{black},
        arc = 2pt,
        rounded corners,
        colframe = black,
        colbacktitle = white!97!yellow,
        colback = white!97!yellow,
        title = #1,
}
\newtcolorbox{promptbox}[3][Prompt]{
colback=black!5!white,
arc=5pt, 
boxrule=0.5pt,
fonttitle=\bfseries,
title=#1, 
before upper={\small}, fontupper=\fontfamily{ptm}\selectfont,
colframe=#2,
label=#3,
}

\usepackage{amsmath,amsfonts,bm}

\def\eqref#1{equation~\ref{#1}}

\def\1{\bm{1}}

\def\vh{{\bm{h}}}

\def\vx{{\bm{x}}}
\def\vy{{\bm{y}}}

\DeclareMathAlphabet{\mathsfit}{\encodingdefault}{\sfdefault}{m}{sl}
\SetMathAlphabet{\mathsfit}{bold}{\encodingdefault}{\sfdefault}{bx}{n}

\newcommand{\E}{\mathbb{E}}

\newtheorem{theorem}{Theorem}[section]
\newtheorem{lemma}[theorem]{Lemma}

\allowdisplaybreaks[2]

\def\1{\mathbbm{1}}

\usepackage{subcaption}  

\usepackage{hyperref}
\usepackage{url}
\title{Gradient Coupling: The Hidden Barrier to Generalization in Agentic Reinforcement Learning}

\author{%
\textbf{Jingyu Liu}\textsuperscript{1{$\star$}},
\textbf{Xiaopeng Wu}\textsuperscript{2{$\star$}}, 
\textbf{Jingquan Peng}\textsuperscript{2}, 
 \textbf{Kehan Chen}\textsuperscript{2}, 
\textbf{Chuan Yu}\textsuperscript{2}, 
\\ 
\textbf{Lizhong Ding}\textsuperscript{3},
\textbf{Yong Liu}\textsuperscript{1,4,5}$^{\dag}$\\
$^1$ Gaoling School of Artificial Intelligence Renmin University of China, Beijing, China \\
$^2$ Taobao \& Tmall Group of Alibaba~ \\
$^3$  Beijing Institute of Technology, Beijing, China \\
$^4$ Beijing Key Laboratory of Research on Large Models and Intelligent Governance \\
$^5$ Engineering Research Center of Next-Generation Intelligent Search and Recommendation, MOE~ \\
\tt\footnotesize liujy1016@ruc.edu.cn\\
}

\begin{document}
\maketitle
\begin{abstract}

    Reinforcement learning (RL) is a dominant paradigm for training autonomous agents, yet these agents often exhibit poor generalization, failing to adapt to scenarios not seen during training. In this work, we identify a fundamental cause of this brittleness, a phenomenon which we term "gradient coupling." We hypothesize that in complex agentic tasks, the high similarity between distinct states leads to destructive interference between gradients. Specifically, a gradient update that reinforces an optimal action in one state can inadvertently increase the likelihood of a suboptimal action in a similar, yet different, state. To solve this, we propose a novel objective where the actor is trained to simultaneously function as a classifier that separates good and bad actions. This auxiliary pressure compels the model to learn disentangled embeddings for positive and negative actions, which mitigates negative gradient interference and improve the generalization performance. Extensive experiments demonstrate the effectiveness of our method. \footnote{The code is available at \url{https://github.com/somebodyhh1/RL_GCD}}
\end{abstract}

\section{Introduction}

Currently, with the rapid development of large language models \citep{GPT-4,gemini,qwen,deepseek_r1,instruction_following}, building an autonomous agent capable of solving real world, complex and long-horizon tasks has gained significant attention given the great power of large language models \citep{agent_first,agent_smart,world_device_control_agent,CodeAgent,empathetic_agents,gui_agent}. However supervised fine-tuning shows unsatisfying performance and can easily cause catastrophic forgetting \citep{RL_generalize} when faced with complex agentic tasks. Therefore, reinforcement learning is now the most effective method to solve multi-turn interaction especially long horizon tasks which require lots of steps to solve \citep{GiGPO,agentevolver,multi_agent_rl}.

Outcome-based reinforcement learning methods, such as GRPO \citep{GRPO,DAPO,reinforce++,logic_rl}, have shown promising performance in domains like mathematical reasoning. Yet, we find that, these methods underperform in multi-turn interactive agentic tasks. Reinforcement Learning can greatly improve the performance on trained environments but the agent tends to perform poorly on unseen environments. If we modify the trained environment, then the model might completely fail to do the job and result in unreasonable actions. For example,
an agent is trained on tasks requiring it to pick up a single object and place it on a table. During testing, if the task is modified to "pick up two objects and place them on the table," the agent—over-reliant on its learned policy—might pick up only the first object and then terminate the task. This action, while optimal within the context of its training data, constitutes a failure in the new, more complex test scenario. This example highlights how policies that are effective in a narrow training setting can be brittle and fail to adapt to novel task requirements.

This raises a critical question: why does Reinforcement Learning exhibit such poor generalization in agentic tasks? We hypothesize that the root cause lies in the inherent high similarity of data within these tasks, which makes gradient descent on one sample also influences the probability of other samples.

In sequential agentic tasks, the input at step $t+1$ is often only a minor variation of the input at step $t$, and different trajectories are also quite similar. This, combined with a limited discrete action space, results in state-action pairs $(s,a)$ that are highly similar across different samples. Consequently, during the training process, the gradient update for one sample can adversely affect the learning of another. For instance, the gradient that correctly reinforces an optimal action $(s_i,a_i)$ might inadvertently strengthen the policy for a similar yet suboptimal pair $(s_j,a_j)$, where $s_i \approx s_j$, $a_i \approx a_j$. We refer to this interaction as "gradient coupling."




The gradient coupling that arises from sample similarity is a double-edged sword. While it can facilitate generalization by allowing the model to apply learned knowledge to novel but similar situations, it becomes harmful when similar observations demand distinct actions. The core challenge, therefore, is not to eliminate gradient coupling entirely, but to selectively control its effects: encouraging positive transfer between appropriate samples while preventing negative interference between good and bad actions.

However, we find that for agentic tasks, the negative interference is quite strong, reinforcing a good action can easily reinforce other bad actions. This is mainly because the representation is similar, so the gradient direction is similar, causing severe gradient coupling. And during reinforcement learning, the learning process only try to strength good actions and punish bad actions, which fails to relieve the negative interference caused by gradient coupling, causing unsatisfying generalization performance.

For more fine-grained control on the gradient coupling, we posit that the model must learn to distinguish between high-quality and flawed actions at the representational level instead of simply reinforcing good actions. In other words, the internal embeddings of an optimal action and a similar yet suboptimal action must be pushed apart.
To achieve this, we propose a simple yet effective solution: reformulating the training objective to compel the agent to simultaneously function as a classifier, explicitly discriminating between positive and negative actions.
Our approach materializes this concept through an auxiliary classification objective. Alongside the primary policy learning task, we train the agent to explicitly classify actions as either "positive" (optimal) or "negative" (suboptimal). This dual-objective forces the model to learn more discriminative representations. Our extensive experiments validate that this integrated approach effectively enhances generalization and task performance.

Our contributions are listed as follows: 
\begin{itemize}
    \item We diagnose the generalization problem in agentic tasks, and attributing it to gradient interference from sample similarity.
    \item We provide a detailed analysis of how gradient interference impacts the training process, we demonstrate how this phenomenon can hinder optimization and generalization.
    \item We propose a novel training paradigm where the agent concurrently learns to effectively decouples harmful gradients, enhances the model's discriminative ability, and significantly improves performance. 
\end{itemize}

\begin{figure*}
  \centering
  \includegraphics[width=\linewidth]{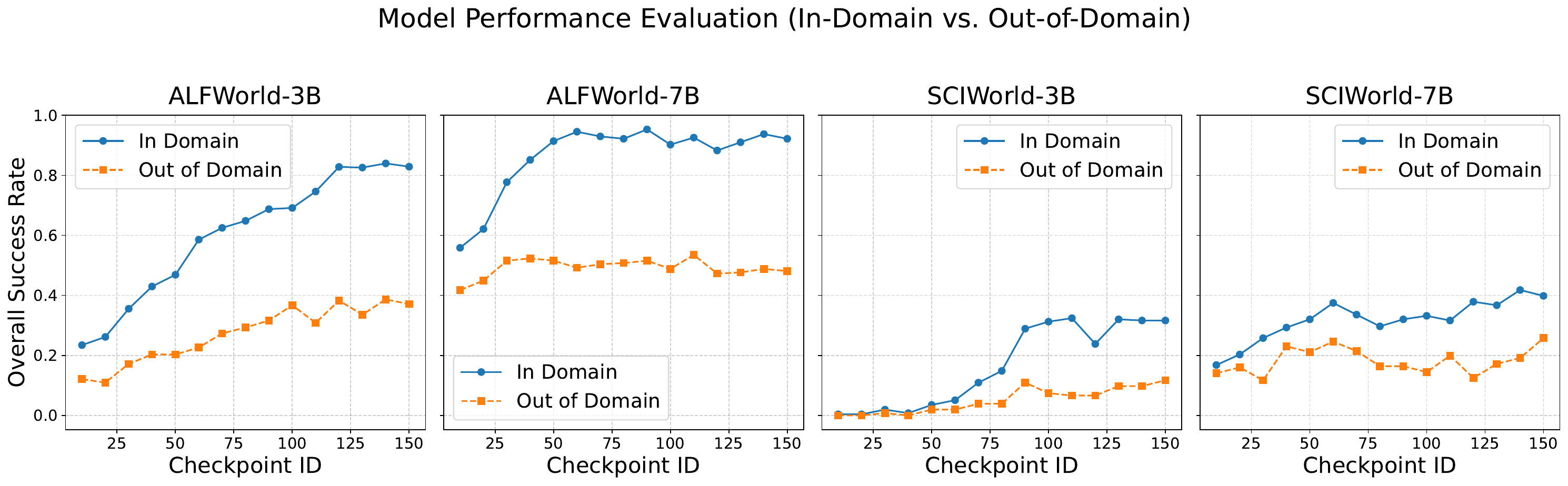}
  \caption{The performance on in domain and out of domain environments}
  \label{fig:performance_instruct}
\end{figure*}

\section{Related Work}
\textbf{Reinforcement Learning for LLM} Reinforcement Learning algorithms like PPO \citep{PPO} is growing extremely popular because it can greatly improve the performance \citep{instruction_following}. Then Direct Preference Optimization (DPO) \citep{DPO} is proposed to simplify the optimization process of PPO, and GRPO removes the critic in PPO and greatly enhances the reasoning performance of LLM by estimating advantages by using batches of samples generated from the same prompt which is also widely applied in tasks like mathematical reasoning \citep{GRPO}, retrieval \citep{atom_searcher,do_not_abstain,search-R1} and tool use \citep{toolRL,CodeAgent} and multi-turn agent tasks \citep{RAGEN,long_horizon_RL_apple,RLVMR,harnessing_uncertainty,sweetRL}. However, we find that although reinforcement learning can greatly help the agent to perform good in the trained task, it might fail to generalize to similar but unseen tasks.

\textbf{Agent Generalization} The agent trained with RL shows great performance on in domain tasks, but the generalization is poor. \citet{SFT_memorize_RL_generalize} claims that RL shows better generalization than SFT, but \citet{rl_no_gene} argue that RL fails to improve the generalization of LLM. \citet{early_experimence} use self reflection and future prediction to improve the generalization. While \citet{agentevolver} automatically collect online data and collect data to further fine-tune the model, improving the performance. And \citet{RLVMR} try to design meta thinking prompts to improve the generalization. And some researchers try to use multi agent interaction to improve the generalization \citep{multi_agent_rl,rema_multi_agent}
But they did not reveal why current methods fail to generalize, only try to solve it through prompt design and data collection while our method try to reveal why agents fail to generalize and solve it.



\section{Gradient Coupling in Agentic Task}

In agentic RL, a critical question is whether Reinforcement Learning (RL) enables agents to acquire generalizable problem-solving skills. To investigate this, we train agents on a specific subset of subtasks using RL. Subsequently, we evaluate their performance on a disjoint set of unseen subtasks. The task split can be seen in Appendix \ref{app:more_experiments}. As shown in Figure \ref{fig:performance_instruct}, the results reveal that the agents' success rates on the unseen subtasks were significantly lower than on the trained subtasks. It shows that standard RL approaches fail to learn the inherent nature of the task it faces difficulty to generalize to unseen tasks.

\begin{figure}
  \centering
  \includegraphics[width=\linewidth]{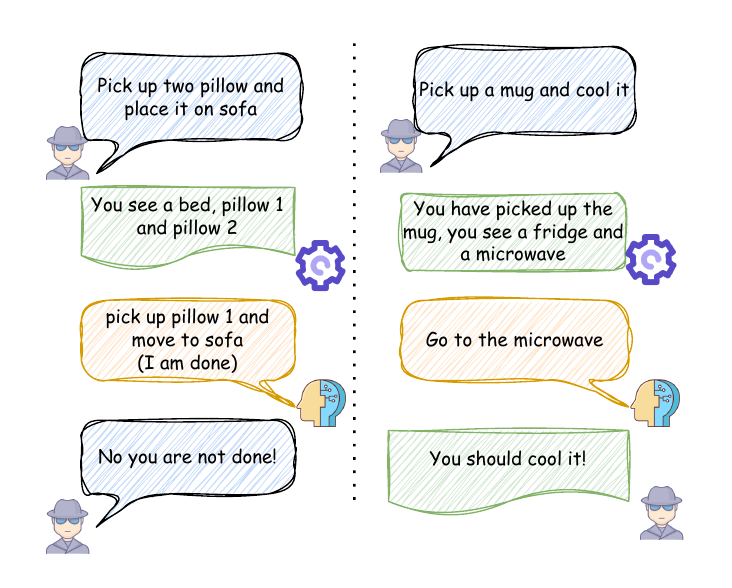}
  \caption{Case study of how the agent fails to generalize}
  \label{fig:case_study}
\end{figure}


We provide concrete examples of these failure modes in Figure \ref{fig:case_study}. For instance, consider an agent trained on the task "pick up an apple and heat it in the microwave." When tested on a novel task, "pick up an apple and cool it in the fridge" the agent, after picking up the apple, incorrectly navigates towards the microwave instead of the fridge. This indicates that the agent has formed a spurious correlation between "picking up an object" and "interacting with the microwave."
Similarly, an agent trained exclusively on "pick up one object and place it on the table" fails when presented with the task "pick up two objects and place them." After successfully retrieving the first object, the agent prematurely proceeds to the destination, completely ignoring the instruction to pick up the second one.

These cases demonstrate that the agent does not learn the underlying semantics of the commands. Instead, it overfits to the specific action sequences seen during training, failing to generalize to variations in task structure or goals.

\subsection{The Influence of Similar Sample Gradients}


Why RL fails to generalize to unseen tasks?
The poor generalization of RL in agentic tasks can be traced back to a fundamental characteristic of the data they generate: the high similarity. Even for different tasks, many of them share similar input observations and otuput actions which frequently leads to trajectories where state-action pairs, or even entire reasoning "thoughts" similar.

\begin{figure}
    \centering
      \includegraphics[width=1\linewidth]{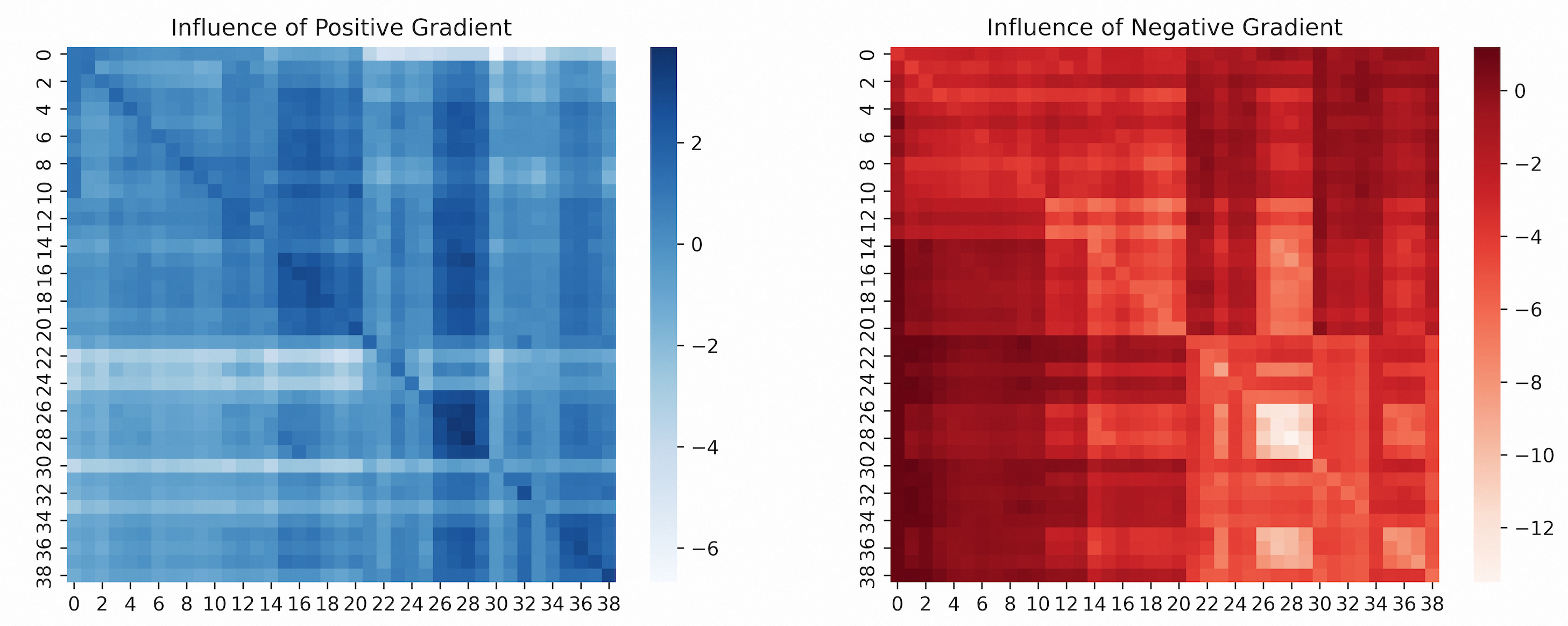}
      \caption{The x axis and y axis stands for different actions (available actions shown in the instruction), and the value is how much the log probability of the action increases/decreases, this directly shows that gradient descent on one sample would influence the probability of other actions}
      \label{fig:influence_of_probability}
\end{figure}

As a result, performing gradient descent on one sample can inadvertently influence the likelihood of other, similar samples. To empirically validate this effect, we conduct experiments on ALFWorld. First, we generate a set of interaction trajectories with the environment. From these, we select pairs of input prompts and outputs, $(x_i, a_i)$ and $(x_j, a_j)$. We then perform a single step of gradient descent using $(x_i, a_i)$ and measure the change in the model’s output probability for the paired sample $(x_j, a_j)$. The results, shown in Figure~\ref{fig:influence_of_probability}, demonstrate a noticeable shift in probability, indicating that optimization on one sample generalizes to similar samples due to shared gradient directions.

\begin{figure*}
  \centering
  \includegraphics[width=\linewidth]{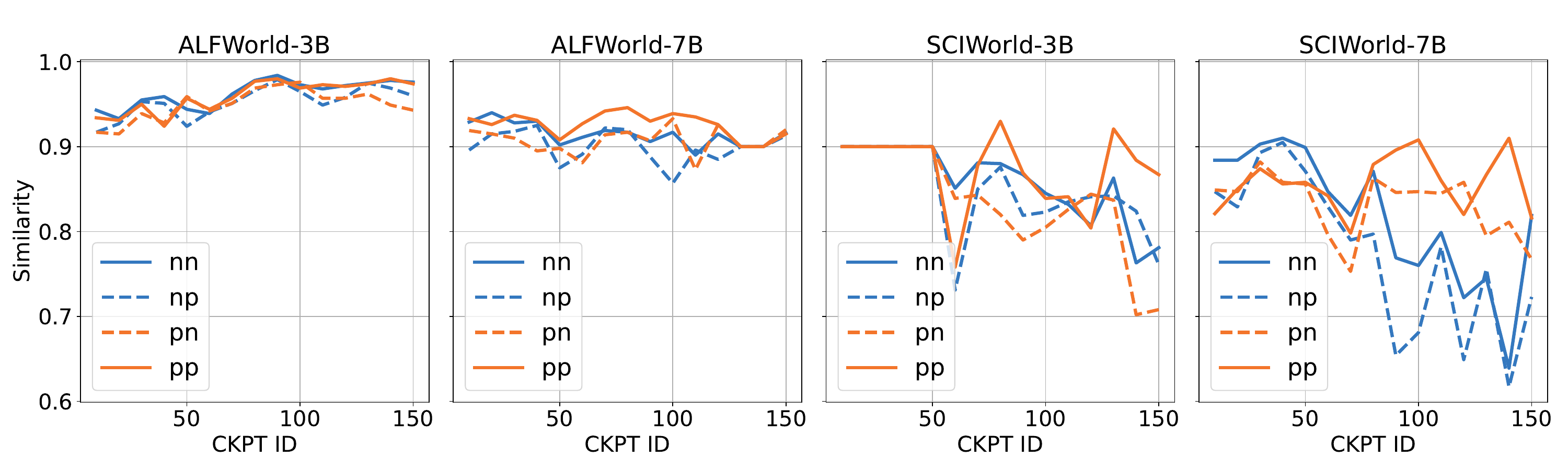}
  \caption{The similarity between gradients of in domain and out domain samples. `nn' denotes a pair of two negative samples, `np' means the similarity between negative and positive. SCIWorld-3B starts with ckpt 50 because the success rate is low at the beginning}
  \label{fig:gradient_influence_change}
\end{figure*}

This implies that the probability of a given action is not only determined by its own gradient signal but is also significantly influenced by gradients from other similar training samples, and we term this "gradient coupling". 
This mechanism is a double-edged sword. On one hand, it can be beneficial, helping the model generalize from one situation to another when similar actions are required. On the other hand, it becomes detrimental when two situations appear similar but demand completely different actions—a common scenario when generalizing from in-domain training data to out-of-domain tasks.

For effective generalization, the model must learn to associate good actions with other good actions, while disassociating them from bad actions. In other words, we want the gradient coupling between successful actions to be high, and the coupling between successful and unsuccessful actions to be low. This would ensure that reinforcing a good action would not mistakenly encourage some bad actions.

\begin{table}[htbp]
  \centering

    \begin{tabular}{cccccc}
    \toprule
          &       & nn    & np    & pn    & pp \\
    \midrule
    \multirow{3}[2]{*}{7B} & MATH  & 0.27  & 0.19  & 0.17  & 0.26  \\
          & ALFWorld & 0.91  & 0.90  & 0.87  & 0.89  \\
          & SCIWorld & 0.80  & 0.83  & 0.74  & 0.85  \\
    \bottomrule
    \end{tabular}%
  \caption{Cosine similarity of gradients for in-domain (ID) and out-of-domain (OOD) samples. `'pp', `nn', and `np' refer to pairs of (positive, positive), (negative, negative), and (negative, positive) samples, respectively. The distinction between `np' and `pn' arises from which sample is drawn from the ID or OOD pool.}
  \label{tab:similarity}%
\end{table}%

To quantify the severity of gradient coupling, we measure the cosine similarity between the average gradients of different sample populations. For two distinct sets of samples, we compute a representative gradient vector across all samples within that set. The cosine similarity is then calculated between these two gradient vectors. A high similarity value indicates strong gradient coupling, as it implies that the parameter update direction for one set is highly aligned with that of the other. In effect, performing gradient descent on one set yields an optimization step that is directionally similar to performing it on the other.

For agentic tasks (ALFWorld, SCIWorld), we instruct the model to interact with the environment with CoT and generate interaction trajectory then compute the average gradient similarity of model parameters between successful (positive) and unsuccessful (negative) samples from both in-domain (ID) and out-of-domain (OOD) data. We perform a similar analysis on mathematical reasoning tasks for comparison, we use MATH and we select 3 type of questions as out domain data, and others as in domain data to calculate the gradient similarity, the detail split is shown in Appendix \ref{app:more_experiments}.

As shown in Table 1, the gradient similarities in ALFWorld and SCIWorld are substantially higher across all sample pairs (e.g., positive-positive, positive-negative) than those observed in the mathematical tasks. This indicates that agentic tasks suffer from far more severe gradient coupling, and the similarity between positive and negative samples are also high, which helps explain their poor generalization.


During reinforcement learning, the agent gradually learns how to do the task and increase the performance on the trained task, then will the gradient coupling be relieved during RL? As we show in Figure \ref{fig:gradient_influence_change}, during GRPO training, the gradient coupling between good and bad actions does not decrease during the training process. This shows that RL does not relieve the gradient coupling between good and bad actions, leading to unsatisfying generalization performance.





\subsection{Gradient Coupling Also Influences in Domain Performance}

Beyond impairing generalization, gradient coupling can also disrupt the optimization process even on the in domain data. Intuitively, a reinforcement learning algorithm should penalize flawed actions that increase the risk of failure while reinforce good actions that lead to success. However, considering the gradient coupling between different samples, a flawed action might be mistakenly influenced by positive advantage samples, and the flawed action could be reinforced instead of punished.

Consider a simple case that the agent is at step $i$, and with current input, the agent might take various of actions, for an action $a_i$ that introduces an additional risk of failure $r>0$ (more chance leading to failure). Let the policy $\pi_{\theta}$ select this action with probability $q$. We can demonstrate that the expected advantage of this action is inherently negative, creating a natural self-correction mechanism.

\begin{lemma} \label{lemma_advantage}
    For a policy $\pi_{\theta}$ with probability $q$ of conducting action $a_i$, conducting action $a_i$ brings risk $r$, then the expected advantage for action $a_i$ is 
    \begin{equation} \label{eqn:advantage}
        \E_{\pi_{\theta}} A_{i}=qr \cdot (q-1),
    \end{equation}
    where $\E_{\pi_{\theta}} A_{i}$ stands for the expected advantage of action $a_i$ (we remove the std for convenience as we only care about the sign of advantage).
\end{lemma}

However, this self-correction can be overpowered by gradient coupling. Consider a flawed action that receives an erroneous "push" from similar but correct samples. This creates a conflict between the internal self-correction signal and the external push from gradient coupling. The outcome of this conflict might mistakenly increase the probability of the flawed action.

Also, the strength of the self-correction mechanism, is not constant. As illustrated in Equation \ref{eqn:advantage}, it follows a parabolic curve, peaking at $q = 0.5$ and decreases with $q$ when $q > 0.5$, where $q$ is the probability of the action. This creates two distinct regimes as follows:

The Safe Regime ($q < 0.5$): The self-correction penalty grows stronger if q increases. This provides a stable negative feedback loop that effectively resists the external push from gradient coupling.

The Danger Zone ($q > 0.5$): As $q$ increases further, the self-correction penalty weakens. This makes the action highly vulnerable to the external push, which can easily overwhelm the feeble self-correction and lead to a runaway increase in the flawed action's probability, cementing it into the policy.

\section{Methodology}



Our analysis shows that gradient coupling could have negatively interference on agentic RL. The core objective, therefore, is to decouple the learning signals for positive and negative trajectories, ensuring that correct actions are reinforced while incorrect ones are suppressed.
While prior work like \citet{GRPO_squeeze} has attempted to mitigate such interference, its methodology is ill-suited for agentic tasks and they does not consider that the gradient coupling can somehow help generalization by connecting positive samples. They try to downweight penalties on specific "harmful" tokens within negative samples to reduce gradient coupling, this strategy is effective in domains like mathematical reasoning, where harmful similarity often originates from generic logical transition words (e.g., "so," "then"). Penalizing these tokens has minimal impact on the core reasoning process. 

In contrast, similarity in agentic tasks stems from more fundamental sources: overlapping action sequences and shared high-level reasoning structures (e.g., planning steps or environmental interactions), as illustrated in Figure \ref{fig:token_influence_gradient}. Penalizing these tokens would hinder the effect of the negative gradient, which could lead unsatisfying performance as we show in Table \ref{tab:performance}.


As shown in \citet{GRPO_squeeze}, the gradient coupling in GRPO can be expressed as:

\begin{equation} \label{eqn:token_influence}
    \sum_{k=1}^{|\vy_i^+|} \sum_{k'=1}^{|\vy_j^-|} \alpha_{k,k'} \cdot \langle \vh_{\vx,y_{i,<k}^+}, \vh_{\vx,y_{j,<k'}^-} \rangle,
\end{equation}
where $\alpha_{k,k'}$ denotes a token-level similarity weight based on prediction error, and $\vh_{\vx,y_{<k}}$ represents the hidden embedding at position $k$ conditioned on input $\vx$ and preceding tokens. This formulation reveals that gradient interference between samples arises primarily from similarities in their internal representations. Therefore, to mitigate this undesirable interaction, we argue that it is essential to disentangle the embeddings of positive and negative samples during training.



This analysis directly motivates our core proposal: Generative Classification Disentanglement (GCD). The central idea is to compel the actor model to simultaneously function as a classifier, thereby forcing it to learn representations that are sensitive to action quality. We implement this by introducing an auxiliary classification objective. 
We formulate $L_{GCD}$ as a secondary reinforcement learning loss calculated using a GRPO-style loss. This process unfolds as follows:

\textbf{Generative Judgment:} We re-purpose the actor model as a generative judge. Given the state $s_i$ and the action $a_i$, we prompt the model to generate a textual judgment, $j_i$, on whether the action was good or bad. To enable a GRPO-style comparison, we generate a group of $N$ such judgments for the same $(s_i, a_i)$ pair, creating a set of responses $\{j_i^{(1)}, j_i^{(2)}, \dots, j_i^{(N)}\}$.

\textbf{Reward Allocation:} Each generated judgment $j_i^{(k)}$ is assigned a reward, $r_{\text{GCD}}^{(k)}$. This reward is simply 1 if the sentiment of the judgment $j_i^{(k)}$ matches the ground-truth label $y_i$, and 0 otherwise. About how to obtain the label of good/bad actions. Inspired by \citet{GiGPO}, if the current observation reoccurs in subsequent steps, this suggests that the action might be incorrect, as it leads to a previously visited state, potentially indicating a loop or redundant move. Otherwise if the current step is the last one, it means that it directly leads to success, so we recognize it as a correct action. To improve data diversity, we also use DeepSeek V3 to generate labels on actions. And we only select those label with high confidence, specifically, we prompt the model for 5 times, if it answer consistently for 4 or more times, then it is used as a training sample.
Then we prompt the actor model to judge if the action is good one and allocate reward to conduct GRPO training.

\begin{table*}[htbp]
  \centering

    \begin{tabular}{cccccccccc}
    \toprule
          &       & \multicolumn{4}{c}{ALFWorld}  & \multicolumn{4}{c}{SCIWorld} \\
\cmidrule{3-10}          &       & L0    & L1    & L2    & mean  & L0    & L1    & L2    & mean \\
    \midrule
    \multirow{7}[2]{*}{Qwen2.5-3B} & vanilla & 16.4  & 15.6  & 9.4   & 13.8  & 0.4   & 0.0   & 0.0   & 0.1  \\
          & reflection & 16.4  & 12.5  & 8.2   & 12.4  & 3.9   & 4.3   & 0.8   & 3.0  \\
          & RLOO  & 81.4  & 68.2  & 40.1  & 63.2  & 28.9  & 27.3  & 13.0  & 23.1  \\
          & GRPO  & 82.7  & 69.4  & 39.7  & 63.9  & 33.2  & 30.5  & 14.2  & 25.9  \\
          & NTHR  & 80.1  & 64.3  & 36.7  & 60.4  & 32.8  & 32.4  & 12.1  & 25.8  \\
          & GCD   & 85.9  & 75.9  & 44.7  & 68.8  & 41.4  & 40.2  & 16.8  & 32.8  \\
          & GCD\_sugg & \textbf{89.3 } & \textbf{78.4 } & \textbf{45.2 } & \textbf{71.0 } & \textbf{42.3 } & \textbf{40.5 } & \textbf{20.1 } & \textbf{34.3 } \\
    \midrule
    \multirow{7}[2]{*}{Qwen2.5-7B} & vanilla & 34.8  & 33.6  & 20.7  & 29.7  & 15.6  & 10.9  & 6.3   & 10.9  \\
          & reflection & 39.8  & 45.7  & 32.8  & 39.5  & 14.5  & 10.5  & 9.8   & 11.6  \\
          & RLOO  & 92.3  & 88.4  & 52.7  & 77.8  & 41.2  & 36.8  & 27.0  & 35.0  \\
          & GRPO  & 92.9  & 87.5  & 48.1  & 76.2  & 41.9  & 36.9  & 27.7  & 35.5  \\
          & NTHR  & 89.4  & 85.7  & 49.5  & 74.9  & 42.3  & 35.2  & 26.5  & 34.7  \\
          & GCD   & 92.7  & \textbf{91.2 } & 57.2  & 80.4  & 42.6  & \textbf{37.1 } & 32.3  & 37.3  \\
          & GCD\_sugg & \textbf{93.2 } & 90.4  & \textbf{61.5 } & \textbf{81.7 } & \textbf{43.4 } & 36.7  & \textbf{33.5 } & \textbf{37.9 } \\
    \bottomrule
    \end{tabular}%
  \caption{The performance of GCD, GCD means we add generative classification task, and GCD\_sugg means that we add explicit suggestion to escape the danger zone. L0 means tasks seen during RL, L1 means in domain unseen tasks, and L2 is out domain unseen tasks.}
  \label{tab:performance}%
\end{table*}%

\textbf{GRPO-style Loss Calculation:} Finally, using the set of rewards $\{r_{\text{GCD}}^{(1)}, \dots, r_{\text{GCD}}^{(N)}\}$ for the group of judgments, we compute the advantage for each judgment $j_i^{(k)}$ and apply the standard GRPO loss.

\begin{equation} \label{eq:gcd_loss}
\begin{split}
    \mathcal{L}_{\text{GCD}} &= -\mathbb{E} \left[ \frac{r_{\text{GCD}}^{(k)} - \bar{r}_{\text{GCD}}}{\sigma(r_{\text{GCD}})} \frac{\pi(j_i^{(k)} | s_i, a_i)}{\pi_{old}(j_i^{(k)} | s_i, a_i)}  \right],\\
    \mathcal{L}&=\mathcal{L}_{\text{GRPO}}+\mathcal{L}_{\text{GCD}},
\end{split}
\nonumber
\end{equation}
where $\bar{r}_{\text{GCD}}$ and $\sigma(r_{\text{GCD}})$ are the mean and standard deviation of rewards within the group. This objective trains the model to become an accurate generative classifier, forcing it to learn representations that are inherently sensitive to the quality of an action, thereby achieving the desired representational disentanglement.

As mentioned before, the gradient coupling is most dangerous when a flawed action’s probability is high (the “Danger Zone"). In this regime, the action’s weak self-correction mechanism might fail to punish the flawed action. Therefore, when generating judgment about whether the action is a good choice, we summarize those judgments and formulate them into a concise suggestion about what should be avoided. Then we insert the suggestion into the prompt to drag the probability of specific flawed actions out of the “Danger Zone’ and into the “Safe Regime." Once the probability is low, the natural self-correction mechanism, can effectively take over and continue to suppress the flawed behavior during reinforcement learning. We summarize the suggestions with DeepSeek-V3. Then we insert the suggestion into the prompt and update the suggestion every 10 epochs.

In our context, the primary RL task ($L_{\text{GRPO}}$) trains the agent on \textit{what to do}, which can lead to superficial, brittle policies based on surface-level similarities. The auxiliary GCD task ($L_{\text{GCD}}$), however, forces the model to also learn \textit{why an action is good or bad}. To succeed at both, the model cannot rely on ambiguous representations. The optimization pressure from $L_{\text{GCD}}$ compels the network to explicitly separate the embeddings of positive and negative trajectories in its latent space.

This induced separation is what directly counters gradient coupling. When the embeddings of a good sample ($h_{\text{good}}$) and a similar but bad sample ($h_{\text{bad}}$) are pushed apart, the gradient from the RL objective for the good sample, $\nabla L_{\text{GRPO}}(h_{\text{good}})$, no longer has a significant projection onto the direction of $h_{\text{bad}}$. Consequently, reinforcing the good action no longer inadvertently strengthens the policy for the bad one. In essence, GCD does not alter the RL gradient itself; rather, it reshapes the underlying representational manifold, making the RL optimization process inherently more precise and less prone to interference. While this mechanism is indirect, its efficacy lies in fundamentally improving the quality of the representations upon which the policy is built.

\section{Experiments} \label{sec:experiments}

\begin{figure*}
  \centering
  \includegraphics[width=\linewidth]{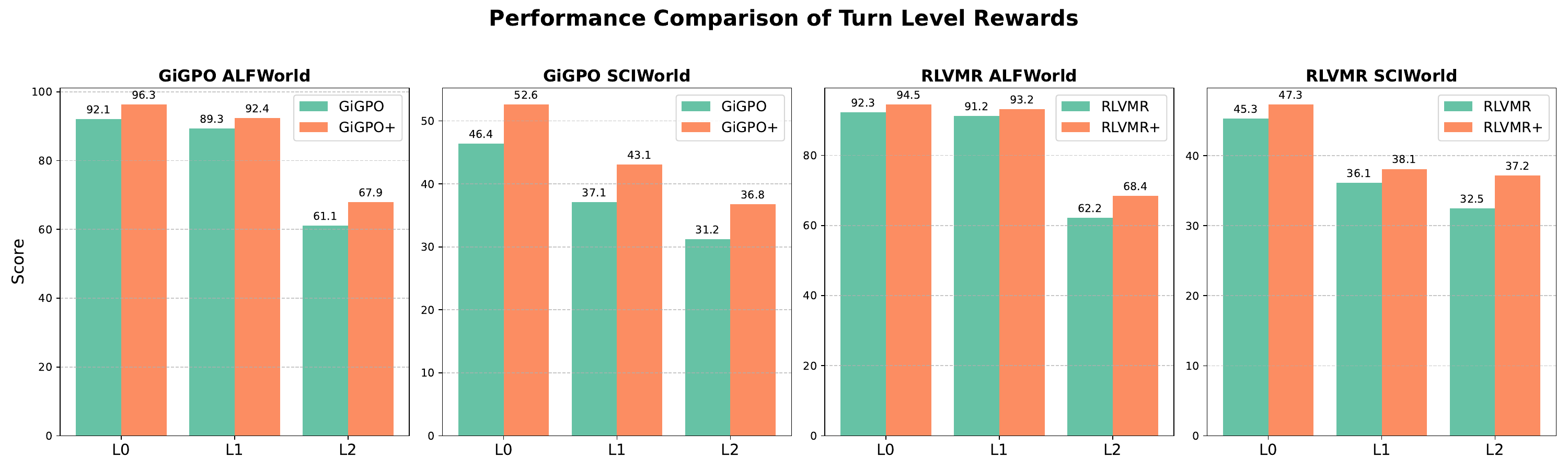}
  \caption{The performance of Qwen2.5-7B-Instruct when our method combined with turn level reward methods. GiGPO+ and RLVMR+ means GiGPO/RLVMR combined with GCD}
  \label{fig:per_step_level_reward}
\end{figure*}

We conduct experiments on ALFWorld \citep{alfworld} and ScienceWorld \citep{agent_smart}, and we train Qwen2.5-3B-Instruct and Qwen2.5-7B-Instruct on the environments fro 150 epochs. We update the suggestion every 10 steps during RL, and we do not add the generated suggestion when testing for fair comparison.  
We only add 30 samples to conduct judgment, so it does not affect the efficiency, and we discuss the time consumption and the performance of the judgment task on in/out domain tasks in Appendix \ref{app:more_experiments}.

We present the results in Table \ref{tab:performance}, the result shows that GCD can help the performance of both in domain tasks and out domain tasks. And we can also observe that NTHR \citep{GRPO_squeeze} who try to downweight penalties on specific "harmful" tokens within negative samples fails to help. As we show in Figure \ref{fig:token_influence_gradient}, the relationship between different samples is caused by lots of meaningful reasoning or actions tokens, penalizing these tokens would hinder the effect of negative gradients, leading to suboptimal performance.

Also, it is worth noting that our method does not help by improving the ability of self reflection because during the reasoning process, the agent does not conduct self reflection during the reasoning process, it directly drives to the action by one thinking without any reflection. This can be observed from Table \ref{tab:response_length} that with GCD, the model responses with similar length, and we show example responses in Appendix \ref{app:more_experiments}

We also conduct experiments with cold start, and the results are shown in Table \ref{tab:per_cold_start}, and we show that if we add some judgment data during the cold start, it also helps the performance.
Also, we show that our method can effectively get rid of the repeated and looped actions in Figure \ref{fig:flaw_action}.

\subsection{The Gradient Coupling Between Samples}
\begin{figure}
  \centering
  \includegraphics[width=\linewidth]{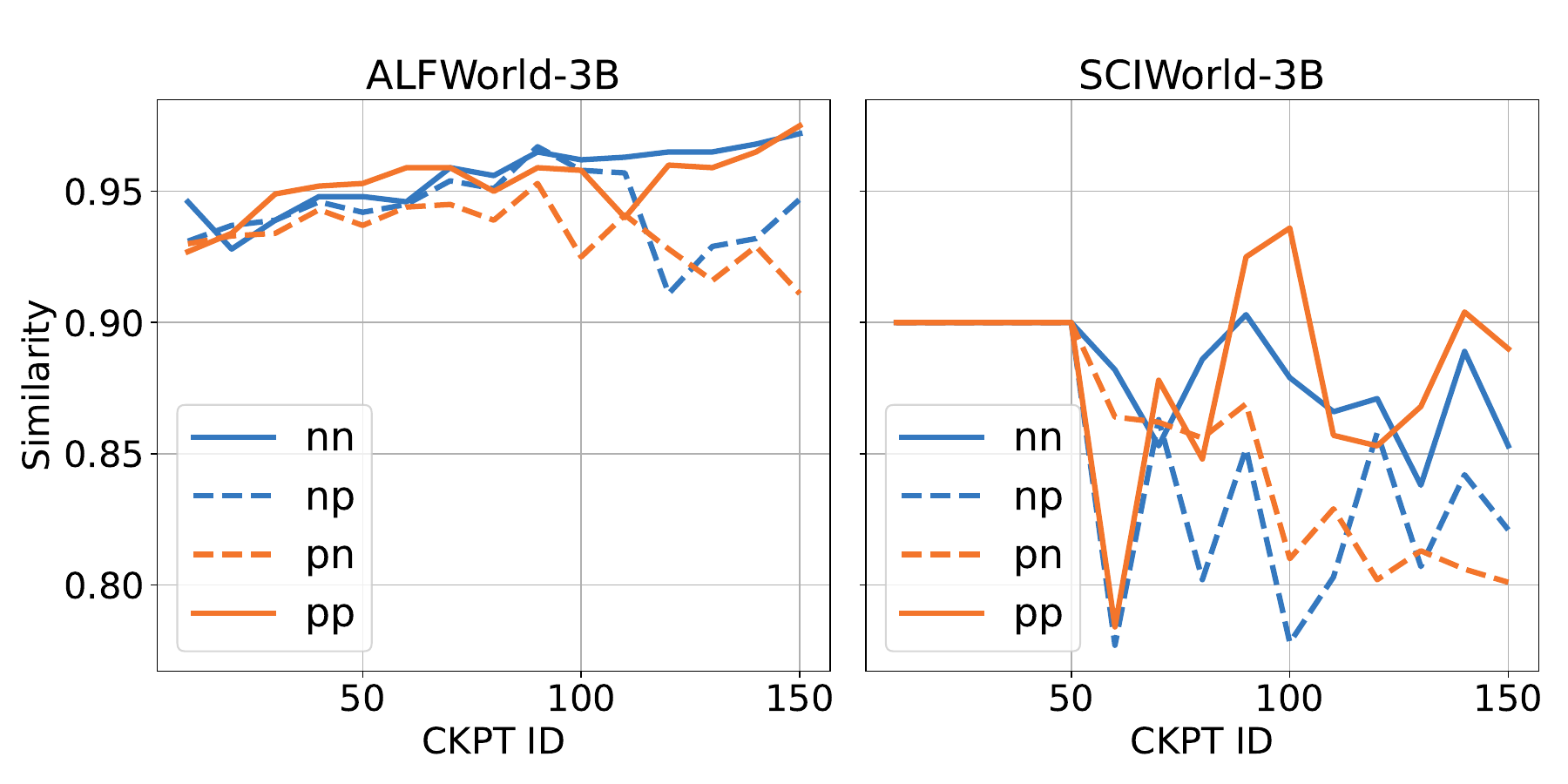}
  \caption{The change of gradient coupling during the training process with GCD. We can observe that np and pn are lower than pp/nn, which leads to better generalization performance. SCIWorld-3B starts with ckpt 50 because the success rate is low at the beginning}
  \label{fig:gradient_influence_change_sub_gcd}
\end{figure}

As we claimed before, our method can effectively separate the embeddings of positive and negative samples, similar with Figure \ref{fig:gradient_influence_change} we calculate the similarity of gradients. The result for Qwen2.5-3B-Instruct is shown in Figure \ref{fig:gradient_influence_change_sub_gcd}, and we show results of Qwen2.5-7B-Instruct in Figure \ref{fig:gradient_influence_change_gcd}. It shows that our method effectively separates the embedding of positive and negative samples, which helps the model to learn helpful actions while getting rid of those harmful ones. By separating the embedding of good and bad actions, our method effectively help both in domain and out domain performance as shown in Table \ref{tab:performance}.

\subsection{Difference with Turn Level Reward}



At first glance, our method's use of turn-level judgments appears related to recent work on turn-level rewards, such as GiGPO \citep{GiGPO} and RLVMR \citep{RLVMR}, which also evaluates intermediate actions. However, we contend that our approach targets a fundamentally different problem. These prior methods primarily address the challenge of sparse rewards and credit assignment. Their goal is to propagate learning signals to crucial intermediate steps within a trajectory, for example, by rewarding beneficial actions even if the overall outcome was a failure.

In contrast, our work addresses the problem of destructive gradient interference between different trajectories. The auxiliary classification task in GCD is not designed to create a more nuanced reward signal. Instead, its sole purpose is to serve as a representational learning objective: forcing the model to distinguish between positive and negative actions at the embedding level. This disentangles their representations and prevents the gradient from a positive sample from incorrectly reinforcing a similar but negative sample.

Given that GCD tackles representational quality while methods like GiGPO tackle credit assignment, they are orthogonal and potentially synergistic. Our experiments in Figure \ref{fig:per_step_level_reward} validate this hypothesis, showing that integrating GCD with these turn-level reward schemes leads to further performance improvements, underscoring the unique contribution of our approach. As GCD is not a reward allocation method, it aims to separate the representation of good and bad actions to reduce the negative interference of gradient coupling, and it can be effectively incorporated into other reward allocation methods and greatly improve the performance.



\section{Conclusion}
This paper addresses the critical problem of poor generalization in agentic reinforcement learning. We identify a fundamental cause as gradient coupling where the high similarity between states in agentic tasks leads to destructive interference during optimization. To solve this, we propose GCD, a novel training objective that distinguishes the representation between good and bad actions which helps the optimization process and generalization. Extensive experiments show its effectiveness.

\section*{Limitations}

The paper discusses why the agent might fail in unseen tasks, and we find that gradient coupling might be the reason, and propose GCD to relieve the negative interference of gradient coupling. However, our method can not fully eliminate the negative interference, and the out domain performance is still significantly lower than in domain performance, which requires further exploration


\bibliography{custom}

\appendix
\newpage

\section{Experiments} \label{app:more_experiments}

\begin{figure*}[htb]
	\centering
	\includegraphics[width=\linewidth]{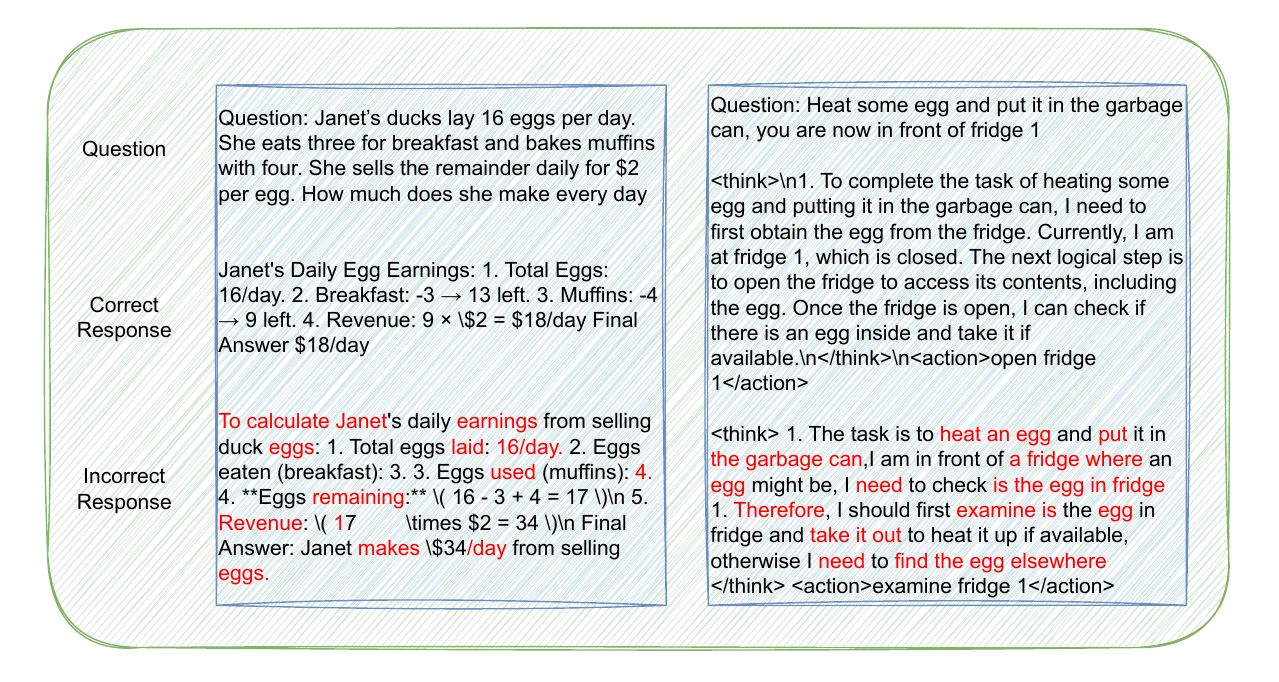}
	\caption{The tokens that cause the similarity in gradient}
	\label{fig:token_influence_gradient}
\end{figure*}

\subsection{Datasets}

We conduct experiments on ALFWorld \citep{alfworld} and ScienceWorld \citep{scienceworld}, ALFWorld is a household environment which requires the agent to explore the room and accomplish tasks like "pick up a pencil and put it on the table". ScienceWorld is designed to evaluate the scientific reasoning abilities with 10 scientific domains and 30 subcategories.
Regarding the in-domain and out-of-domain split, following \citet{RLVMR}, for ALFWorld, we use Cool \& Place and Pick Two \& Place as the out domain task and other 4 tasks as in domain task; for ScienceWorld, the final task type of each topic is reserved as out domain task evaluation.

For MATH 500 dataset, we use Number Theory, Intermediate Algebra, Counting \& Probability and Geometry as in domain data and others as out domain data.

For cold start, we use DeepSeek-V3 to generate trajectories and select the success ones to conduct cold start, then we generate 100 pieces of judgment/future data and incorporate it into the training data and we train the model for 5 epochs. For reinforcement learning, we use train batch size 16 and group size 8 and learning rate 1e-6, the max steps of interaction with environment is set to 30, and for both GCD task and agent task, we sample 8 different responses for one sample.

We show the performance when conducting cold start with different settings in Table \ref{tab:per_cold_start}, we compare between vanilla cold start (GRPO), and adding some data to predict what will happen after certain actions (future GRPO), and adding some judgment data in cold start (judge GRPO). And judge GCD means that we conduct cold start with judgment data and use GCD as auxiliary task during reinforcement learning. We can observe that by conducting cold start with some generative judge data, the performance increases.


\subsection{Flawed Actions}

\begin{figure*}
  \centering
  \includegraphics[width=\linewidth]{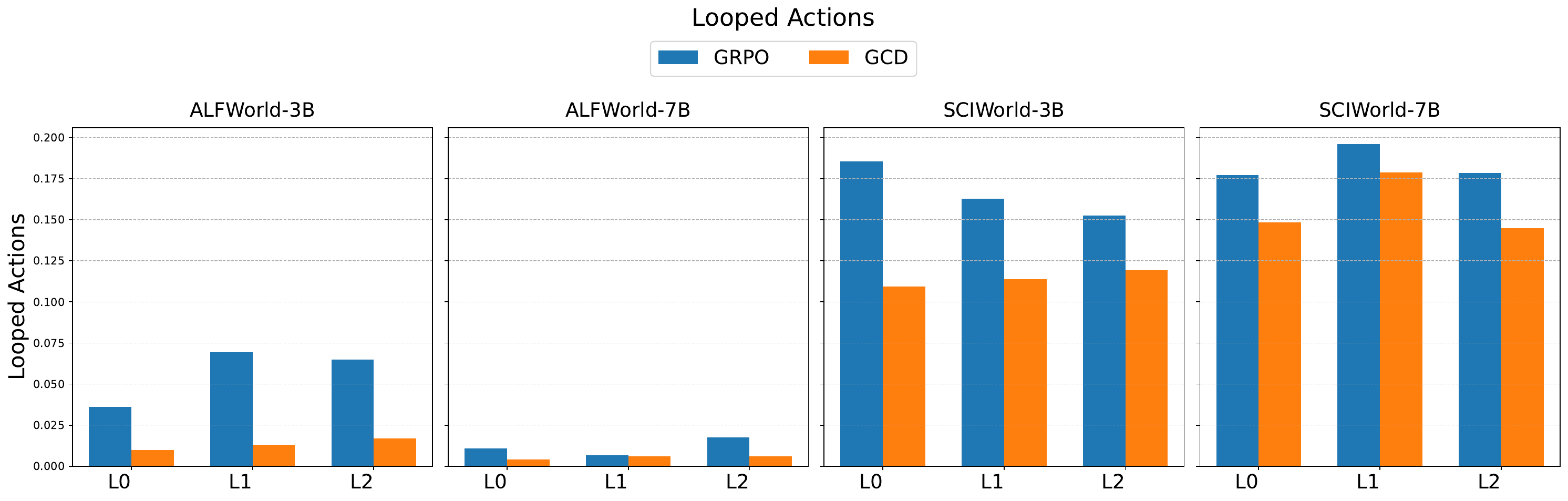}
  \caption{Percentage of Repeated or Looped actions.}
  \label{fig:flaw_action}
\end{figure*}

We observe that the model might conduct repeated actions, which means it directly copy the action of the previous step, or it might be stuck in a loop, which means it keeps repeating some sequence of actions. We can find that our method can effectively reduce the number of those bad actions, with GCD, the model conducts fewer repeated actions as shown in Figure \ref{fig:flaw_action}.



\subsection{Time Consumption} \label{app:time_consumption}



\begin{table}[htbp]
  \centering
  
    \begin{tabular}{cccc}
    \toprule
          &       & ALFWorld & SCIWorld \\
    \midrule
    \multirow{2}[2]{*}{Qwen2.5-3B} & GRPO  & 19.24 & 17.29 \\
          & GCD   & 21.05 & 19.21 \\
    \midrule
    \multirow{2}[2]{*}{Qwen2.5-7B} & GRPO  & 25.09 & 26.81 \\
          & GCD   & 25.99 & 27.87 \\
    \bottomrule
    \end{tabular}%
    \caption{The time consumption (hours) of different methods.}
  \label{tab:time_cost}%
\end{table}%

We can observe from Table \ref{tab:time_cost} that our method does not add too much time consumption, this is mainly because we only select 30 samples for the agent for judgment, and the number is much smaller than the interaction with the environment, so the time consumption is close.

\subsection{Judgment Accuracy}

We also show the accuracy of judgment on L0, L1 and L2 tasks in Table \ref{tab:judge_acc}, we can observe that the agent can effectively judge if a action is good or not. Also the judgment shows great generalization, it shows good performance on L2 tasks, we assume this is mainly because, although the tasks are different, but the mistakes taken is similar.

\begin{table}[htbp]
  \centering

    \begin{tabular}{ccccccc}
    \toprule
          & \multicolumn{3}{c}{ALFWorld} & \multicolumn{3}{c}{SCIWorld} \\
          & L0    & L1    & L2    & L0    & L1    & L2 \\
    \midrule
    Qwen2.5-3B & 65.6  & 63.5  & 59.4  & 75.2  & 74.3  & 67.1 \\
    Qwen2.5-7B & 69.4  & 65.3  & 62.1  & 85.2  & 83.7  & 74.2 \\
    \bottomrule
    \end{tabular}%
  \caption{The performance when we instruct the agent to conduct judgment after GCD training, it shows that the agent can also generate valuable judgment on unseen tasks.}
  \label{tab:judge_acc}%
\end{table}%

\begin{table}[htbp]
  \centering
  
    \begin{tabular}{cccc}
    \toprule
          &       & ALFWorld & SCIWorld \\
    \midrule
    \multirow{2}[2]{*}{Qwen2.5-3B} & GRPO  & 148 & 217 \\
          & GCD   & 136 & 208 \\
    \midrule
    \multirow{2}[2]{*}{Qwen2.5-7B} & GRPO  & 109 & 267 \\
          & GCD   & 116 & 254 \\
    \bottomrule
    \end{tabular}%
    \caption{The Response Length of Our Method}
  \label{tab:response_length}%
\end{table}%

\begin{table*}[htbp]
  \centering
  
    \begin{tabular}{cccccccccc}
    \toprule
          &       & \multicolumn{4}{c}{ALFWorld}  & \multicolumn{4}{c}{SCIWorld} \\
\cmidrule{3-10}          &       & L0    & L1    & L2    & mean  & L0    & L1    & L2    & mean \\
    \midrule
    \multirow{4}[2]{*}{Qwen2.5-3B} & GRPO  & 88.3  & 87.2  & 50.4  & 75.3  & 53.1  & 51.3  & 28.4  & 44.3  \\
          & future\_GRPO & 88.7  & 87.9  & 53.8  & 76.8  & 55.3  & 53.4  & 31.7  & 46.8  \\
          & judge\_GRPO & 89.2  & 87.4  & 56.7  & 77.8  & 53.8  & 53.4  & 33.3  & 46.8  \\
          & judge\_GCD & \textbf{91.2 } & \textbf{90.4 } & \textbf{63.5 } & \textbf{81.7 } & \textbf{55.3 } & \textbf{54.2 } & \textbf{37.9 } & \textbf{49.1 } \\
    \bottomrule
    \end{tabular}%
    \caption{The performance with cold start, GRPO means standard GRPO with cold start, we use 1000 samples and conduct cold start for 5 epochs, future\_GRPO means we add 100 sample about predicting what will happen after certain actions, judge\_GRPO means we add 100 judgment data, and judge\_GCD means we add generative classification task during RL based on the judgment cold start.}
  \label{tab:per_cold_start}%

\end{table*}%

\begin{figure*}
  \centering
  \includegraphics[width=\linewidth]{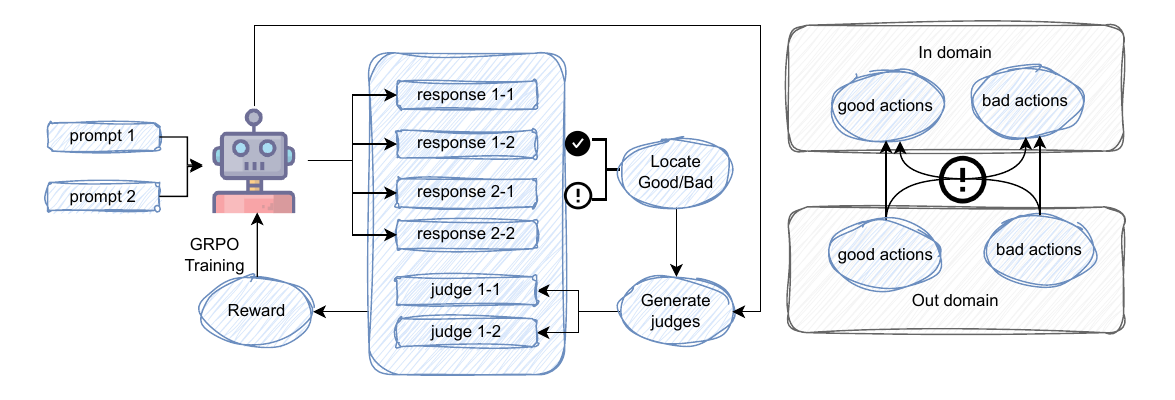}
  \caption{The pipeline of GCD, we sample some actions and instruct the agent to conduct judgment and allocate reward.}
  \label{fig:GRPO_cri_pipeline}
\end{figure*}

\begin{figure*}
  \centering
  \includegraphics[width=\linewidth]{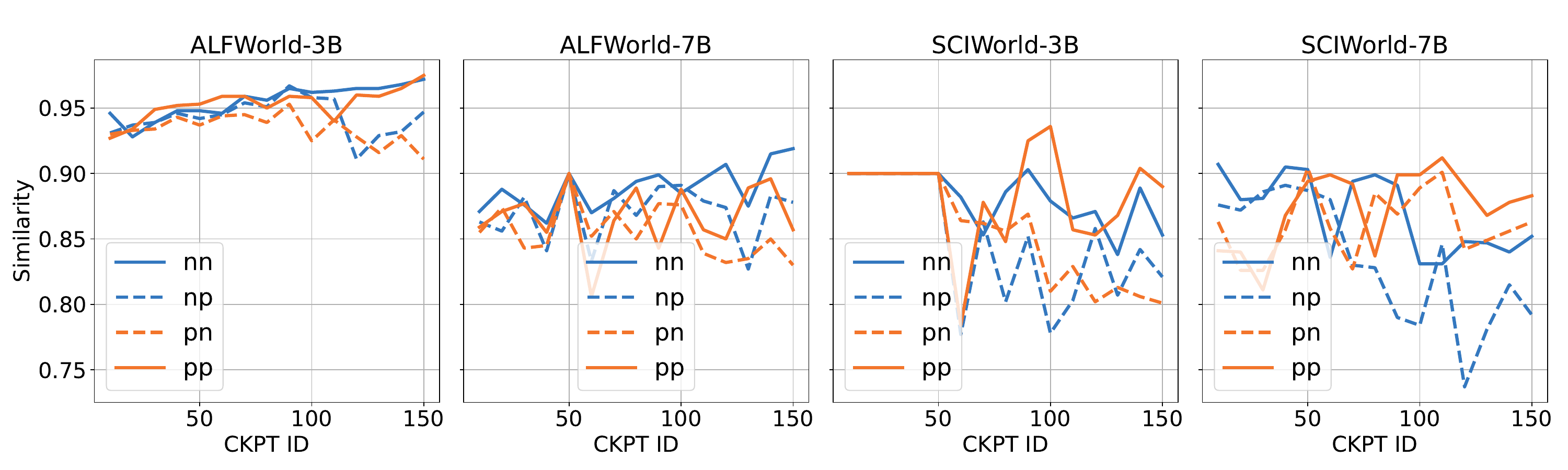}
  \caption{The similarity between gradients of in domain and out domain samples. `nn' denotes a pair of two negative samples, `np' means the similarity between negative and positive.}
  \label{fig:gradient_influence_change_gcd}
\end{figure*}

\begin{tcolorbox}[title = {Response of GCD in ALFWorld for Qwen3B}]
$\langle think \rangle$ Given the task to clean some fork and put it in the drawer, I need to first find a fork. Forks are commonly found on countertops or in the dining area of a kitchen. The countertop 1 is the most likely place to find a fork. After finding a fork, I will need to clean it using the sinkbasin or dishwasher (if available). Since the task does not specify the need for a dishwasher, I will clean the fork in the sinkbasin. Finally, I will put the clean fork in the drawer. The countertop is the best place to find a fork, so I will start by checking there.$\langle /think \rangle$ $\langle action \rangle$ go to countertop 1 $\langle /action \rangle$
\end{tcolorbox}

\begin{tcolorbox}[title = {Response of GCD in ScienceWorld for Qwen3B}]
$\langle think \rangle$ Based on the current observation, we need to determine which inclined plane has the least friction. To do this, we need to test the two inclined planes by using a block on them. However, the given objects do not include a block. We can use the block from the table as it is the only object that can be used to test the friction between the inclined planes and the surface of the planes (wood and chocolate). First, we need to pick up the block from the table, then we can test the friction by placing the block on each inclined plane and observing which one causes the block to slide down more slowly (indicating less friction).$\langle /think \rangle$ $\langle action \rangle$pick up block$\langle /action \rangle$
\end{tcolorbox}

\onecolumn
\section{Proofs} \label{proof_squeeze}

\subsection{Proof of Lemma \ref{lemma_advantage}}

Given a policy $\pi_{\theta}$, with probability $p$ it would fail completing the task and get reward $0$, otherwise success with reward $1$. However, with probability $q$, it will take a mistake, and the mistake could enlarge the risk of failing by $r$.
In this way, the success and fail probability could be represented as follows

\begin{equation}
\left\{\begin{matrix}
  (1-q) \cdot (1-p),& \text{no mistake and success} \\
  (1-q) \cdot p,& \text{no mistake but fail}\\
  q \cdot (1-p-r),& \text{mistake but success} \\
  q \cdot (p+r),& \text{mistake and fail}
\end{matrix}\right.
\end{equation}
Then, we can calculate that,
\begin{equation}
\begin{split}
    \mathbb{P}(success)&=(1-q) \cdot (1-p) + q \cdot (1-p-r)=1-p-qr, \\
    \mathbb{P}(fail)&=(1-q) \cdot p+q \cdot (p+r)=p+qr. \\
\end{split}
\label{eqn:success_fail_prob}
\end{equation}

In GRPO, the advantage is calculated as $A_i=\frac{s_i-\E(s)}{\text{std}(s)}$, where $A_i$ means the advantage, $s_i$ is the calculated reward score. Based on Equation \ref{eqn:success_fail_prob}, we have that $\E (a)=1 \cdot \mathbb{P}(success)+ 0 \cdot \mathbb{P}(fail)=1-p-qr$.
Then, we have that 

\begin{equation}
A_i=\left\{\begin{matrix}
  \frac{p+qr}{\text{std}(r)},& a_i=1\\
  \frac{p+qr-1}{\text{std}(r)},& a_i=0
\end{matrix}\right.
\end{equation}

Then the expected advantage about the mistake is (we ignore the std),

\begin{equation}
\begin{split}
    A_{mistake}
    &=q \cdot (1-p-r) \cdot (p+qr) +  q \cdot (p+r) \cdot (p+qr-1)\\
    &=(q-pq-qr) \cdot (p+qr)+(pq+qr) \cdot (p+qr-1)\\
    &=(q-pq-qr+pq+qr) \cdot (p+qr)-pq-qr\\
    &=pq+q^2r-pq-qr=qr(q-1)
\end{split}
\label{eqn:advantage_mistake}
\end{equation}

This means that unless $q=1$ or $r=0$, the advantage will be negative, which means unless the mistake will happen with probability $1$ or the mistake will not lead to any risk of failure, it will be discouraged.

In contrast, if $r< 0$, which means that the action will reduce the risk of failure, then it will be encouraged unless the probability of taking the action is already $1$.

\section{Prompts}
\label{app:prompts}

\begin{tcolorbox}[title = {Prompt for ALFWorld}]
You are an expert agent operating in the ALFRED Embodied Environment. Your task is to: \{task description\}
You should strictly follow the guidelines below, do what the guideline suggests in your thinking steps to make better actions:
\{suggestions\}
Prior to this step, you have already taken \{step count\} step(s). Below are the most recent \{history length\} observaitons and the corresponding actions you took: \{action history\}
You are now at step \{current step\} and your current observation is: \{current observation\}
Your admissible actions of the current situation are: [\{admissible actions\}].

Now it's your turn to take an action.
You should first reason step-by-step based on the guideline about the current situation. This reasoning process MUST be enclosed within $\langle think \rangle$ $\langle /think \rangle$ tags.   
Once you've finished your reasoning, you should choose an admissible action for current step and present it within $\langle action \rangle$ $\langle /action \rangle$ tags.
\end{tcolorbox}

\begin{tcolorbox}[title = {Prompt for ScienceWorld}]
You are an expert agent operating in the ScienceWorld environment, which is a text-based virtual environment centered around accomplishing tasks from the elementary science curriculum.
You should strictly follow the guidelines below, do what the guideline suggests in your thinking steps to make better actions:
\{suggestions\}
Your current task is: \{task description\}

Prior to this step, you have already taken \{step count\} step(s). Below are the most recent \{history length\} observations and the corresponding actions you took: \{action history\}
You are now at step \{current step\} and your current observation is: \{current observation\}
Here are the actions you may take:

Current available actions:
\{available actions\}

Now it's your turn to take an action. You should first reason step-by-step about the current situation. This reasoning process MUST be enclosed within $\langle think \rangle$ $\langle /think \rangle$ tags.
Once you've finished your reasoning, you should choose an appropriate action for the current step and present it within $\langle action \rangle$ $\langle /action \rangle$ tags.
\end{tcolorbox}

\begin{tcolorbox}[title = {Prompt for generate suggestion}]

\textbf{Role:} You are an Expert AI Strategist, tasked with synthesizing raw feedback to generate high-level, actionable principles for improving an agent's performance. The agent is operating in a complex environment like ALFWorld or ScienceWorld.

\textbf{Context:}
You will be provided with a pre-defined theme and a list of specific advice sentences that have already been clustered under this theme.

\textbf{Your Sole Task:}
Distill the entire collection of related advice into one single, overarching "Golden Rule."

This rule must be:
\begin{itemize}
    \item High-Level: Abstract away from specific examples.
    \item Actionable: Provide clear guidance on what the agent should do.
    \item Generalizable: Be applicable to future, unseen situations related to this theme.
\end{itemize}

For each theme provided in the input data, you must perform the following steps:
1.  Think Step-by-Step: First **you must think step by step and dig deep into the advices to formulate the high-level principle.** Analyze the specific advice, identify the common pattern or root cause, and build a line of reasoning toward a general rule.
2.  Formulate the Rule: Based on your thinking, synthesize the advice into **one single, actionable, and generalizable Golden Rule."

ADVICE LIST:
\{advice list\}
\end{tcolorbox}

\begin{tcolorbox}[title = {Prompt for judgment generation}]
**Role:** You are an expert evaluator, a "Critic" tasked with judging the quality of an action taken by another agent in ALFWorld/SciWorld, a household/science environment where the agent is required with some tasks like 'put a cellphone in bed' or 'determine if metal fork is electrically conductive'

Given the question and the agent response you should judging the quality of the response

**Context:**
*   **Problem:** {problem}
*   **Agent Response:** {response}

**Your Task:**
1.  **Analyze:** In your `$\langle think \rangle$` block, perform a step-by-step analysis.
    *   Consider the `Current State` and the overall `Problem`.
    *   Evaluate will the 'Chosen Action' cause serious trouble or it is obvious not a good option

2.  **Judge:** In the `$\langle answer \rangle$` block, provide a score. Use `1` for a good action, and `0` if the action is obviously a poor action and there exists explicitly better actions.

3.  **Advise:** In the `$\langle advise \rangle$` block, provide a single, concise sentence of feedback.
    *   Based on your analysis of current action and previous actions, provide **advice** to help the agent avoid its mistakes and ehnahce its advantage in the future.
    *   Provide a high-level strategic principle that applies to various situations, rather than a correction for this specific instance.
    
**Output Format:**

$\langle think \rangle $
Your step-by-step reasoning here.
$\langle /think \rangle $
$\langle answer \rangle\\boxed{{0/1}}\langle /answer \rangle $
$\langle advise \rangle$ Your concise, high-level strategic advice here $\langle /advise \rangle$
\end{tcolorbox}

\end{document}